# Homogeneous Spiking Neuromorphic System for Real-World Pattern Recognition

Xinyu Wu, *Student Member, IEEE*, Vishal Saxena, *Member, IEEE*, and Kehan Zhu

***Abstract*—A neuromorphic chip that combines CMOS analog spiking neurons and memristive synapses offers a promising solution to brain-inspired computing, as it can provide massive neural network parallelism and density. Previous hybrid analog CMOS-memristor approaches required extensive CMOS circuitry for training, and thus eliminated most of the density advantages gained by the adoption of memristor synapses. Further, they used different waveforms for pre and post-synaptic spikes that added undesirable circuit overhead. Here we describe a hardware architecture that can feature a large number of memristor synapses to learn real-world patterns. We present a versatile CMOS neuron that combines integrate-and-fire behavior, drives passive memristors and implements competitive learning in a compact circuit module, and enables in-situ plasticity in the memristor synapses. We demonstrate handwritten-digits recognition using the proposed architecture using transistor-level circuit simulations. As the described neuromorphic architecture is homogeneous, it realizes a fundamental building block for large-scale energy-efficient brain-inspired silicon chips that could lead to next-generation cognitive computing.**

## I. Introduction

The human brain is a very energy-efficient computing machine: tasks like perception, object recognition, speech recognition and language translation are trivial to a human brain; whereas modern machines can do such tasks, but require orders of magnitude more energy, as well as specialized programming. Massive parallelism is one of the reasons our brains are so effective in the above mentioned decision-making tasks. Radically different from today's predominant von Neumann computers (memories and processing elements are separated), a biological brain stores memory and computes using similar motifs. Neurons perform computation by propagating spikes and storing memories in the relative strengths of their synapses as well as their interconnectivities. By repeating such a simple structure of neurons and synapses, a biological brain realizes a very energy-efficient computer. Inspired by such architecture, artificial neural networks (ANNs) have been developed and achieved remarkable success in a few specific applications, but historically require hardware resource intensive training methods (such as the gradient-based back-propagation algorithms) on conventional computers, and therefore making them inefficient computationally and in energy use. By exploiting parallel graphical processing units (GPUs) or field programmable gate arrays (FPGAs), power consumption of neural networks has been reduced by several orders of magnitude [1], which yet remains far higher than their biological counterparts.

In the past decade, the discovery of spike-timing-dependent-plasticity (STDP) [2]–[8] has opened new avenues in neural network research. Theoretical studies have suggested STDP can be used to train spiking neural networks (SNNs) in-situ without trading-off their parallelism [9]–[12]. Further, nano-scale memristive devices have demonstrated biologically plausible STDP behavior in several experiments [13]–[17], and therefore have emerged as an ideal candidate for electrical synapses. To this end, hybrid CMOS-memristor analog very-large-scale integrated (VLSI) circuits have been proposed [18]–[22] to achieve dense integration of CMOS neurons and memristors for brain-inspired computing chips by leveraging the contemporary nanometer silicon processing technology.

Researchers have recently demonstrated pattern recognition applications on spiking neuromorphic systems (with memristor synapses) [23]–[32] using leaky integrate-and-fire neurons (IFNs). Most of these systems either require extra training circuitry attached to the synapses (thus eliminating most of the density advantages gained by using memristors), or different waveforms for pre- and post-synaptic spikes (thus introducing undesirable circuit - overhead which significantly limit power and area budget of a large-scale neuromorphic system). There have been a few CMOS IFN designs that attempt to accommodate memristor synapses and in-situ synaptic plasticity together. An asynchronous IFN architecture was proposed in [33], [34], which provided current summing nodes, and propagated same-shape spikes in both the forward and backward directions. Another CMOS IFN with a current conveyor was implemented to drive the memristor as excitatory

This work is supported in part by the National Science Foundation under the Grant CCF-1320987. The work of X. Wu and K. Zhu are supported in part by the graduate fellowship of Boise State University.

The authors are with the Electrical and Computer Engineering Department, Boise State University, Boise, ID 83725, USA (e-mail: xinyuwu@u.boisestate.edu; vishalsaxena@boisestate.edu; kehanzhu@u.boisestate.edu).

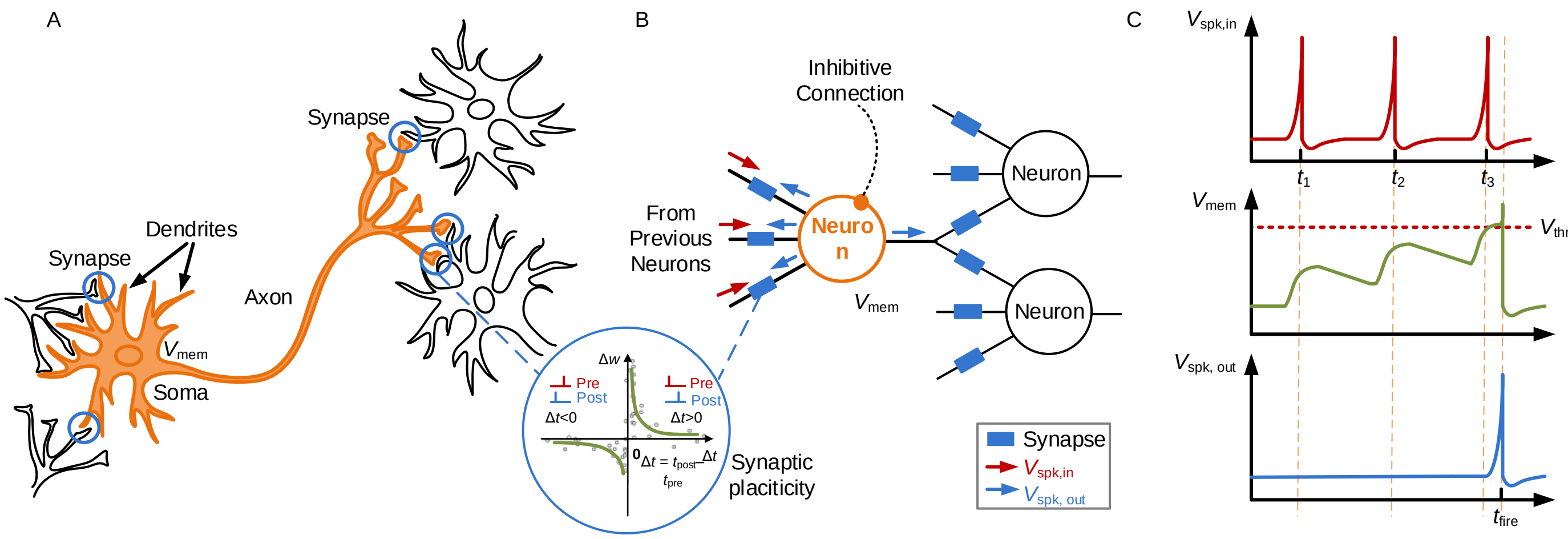

Fig. 1. (A) Simplified diagram of a typical biological neural cell. Soma receives synaptic signals from other neurons through its dendrites, and axon propagates signals to other neurons. A synapse is a contact between the axon of one neuron and a dendrite of another. Soma maintains a voltage gradient across neuron membrane. If the voltage changes by a large enough amount, an action potential pulse, called spike, is generated, then travels along the axon, and eventually activates synaptic connections with other cells when it arrives. (B) A neuromorphic network models the spiking neural network, and (C) Working mechanism of a typical integrate and firing neuron. The neuron maintains membrane voltage $V_{mem}$; once $V_{mem}$ crosses a firing threshold $V_{thr}$, the neuron fires and sends a spike $V_{spk,out}$ to pre and post-synaptic neurons which are connected to it. Synaptic strength, is also called synaptic weight $w$, can be modulated by the pre- and post-synaptic spikes, which is called synaptic plasticity. The experimental example of pair-wise STDP learning curve shown in the circle was redrawn from [3].

or inhibitory synapse [35], [36]. However, none of them supports pattern classification directly owing to the lack of a mechanism for making decisions when employed in a neural network. Moreover, the consideration of large current drive capability for a massive number of passive memristor synapses was absent in these designs.

In this paper, we describe a neuromorphic architecture that can scale to a large number of memristor synapses to learn real-world patterns. To do so, a versatile CMOS spiking IFN was developed. A winner-takes-all (WTA) interface is embedded to empower competitive learning with a shared WTA bus topology among local neurons. A dynamic powering scheme is used to achieve large current drive capability without compromising the energy-efficiency. By exploiting a reconfigurable architecture inspired by [34], the neuron accommodates symmetric forward and backward propagation of spikes for online STDP. With a new tri-mode operation, the neuron encapsulates all functions with a single OpAmp in a very compact circuit, while allowing one-terminal connectivity between the neuron and a synapse. Consequently, it enables a simple repeating homogenous structure with a fully asynchronous communication protocol, and thus facilitates scaling-up to large-scale neuromorphic chips. Employing an industry-standard circuit simulator, we show online STDP learning in memristors and large current drive capability with high energy-efficiency of the proposed neuron, and demonstrate a handwritten-digits recognition application using the proposed architecture.

The rest of this article is organized as follows: Section II introduces the system architecture and building blocks needed to realize a homogeneous neuromorphic system; Section III proposes the CMOS neuron topology and explains how it works as a fundamental information processing unit; Section IV presents a pattern recognition application using the proposed system; Section V demonstrates operations of the proposed CMOS neuron, STDP learning in memristors and an 8×8 handwritten-digits recognition; finally, Section VI discusses the limitations and future challenges.

## II. Homogeneous Neuromorphic System

Fig. 1B shows a basic neuromorphic unit which comprises several synapses and a neuron block. It mimics a biological neuron as shown in Fig. 1A, where the synapse receives spikes from other neurons and converts them into currents according to their synaptic strength. The neuron block performs spatio-temporal integration of the spikes and generates output spikes (or action potentials) similar to the operation of a neuron soma (Fig. 1C). Further, the dendrites and axons are implemented using interconnect circuits which model the spiking-signal propagation through neuronal fibers and used to realize larger signal processing networks [37].

### *A. Memristor as Synapse*

The memristor was first conceptually conceived in 1971 by Leon Chua [30] from a circuit theory perspective. In theory, a memristor is a two-terminal device that can retain an internal analog state by the value of its resistance*, or conductance, that depends upon on the history of the applied voltage and thus the current flowing through the device. Since the conductance of a memristor can be incrementally increased or decreased by controlling the flux through it, it is a potential candidate for realizing electronic equivalent of biological synapses. However, memristor based neural networks have only begun to be explored due to the recent emergence of nano-scale memristor devices.

Memristance has recently been demonstrated in nano-scale

* Memristance, resistance, conductance, synaptic weight and synaptic strength are the different descriptions for the same character of a memristor synapse. For convenience, we use conductance, which is proportional to synaptic weight as used in computer science or synaptic strength as used in neuroscience, when we refer to memristor device.

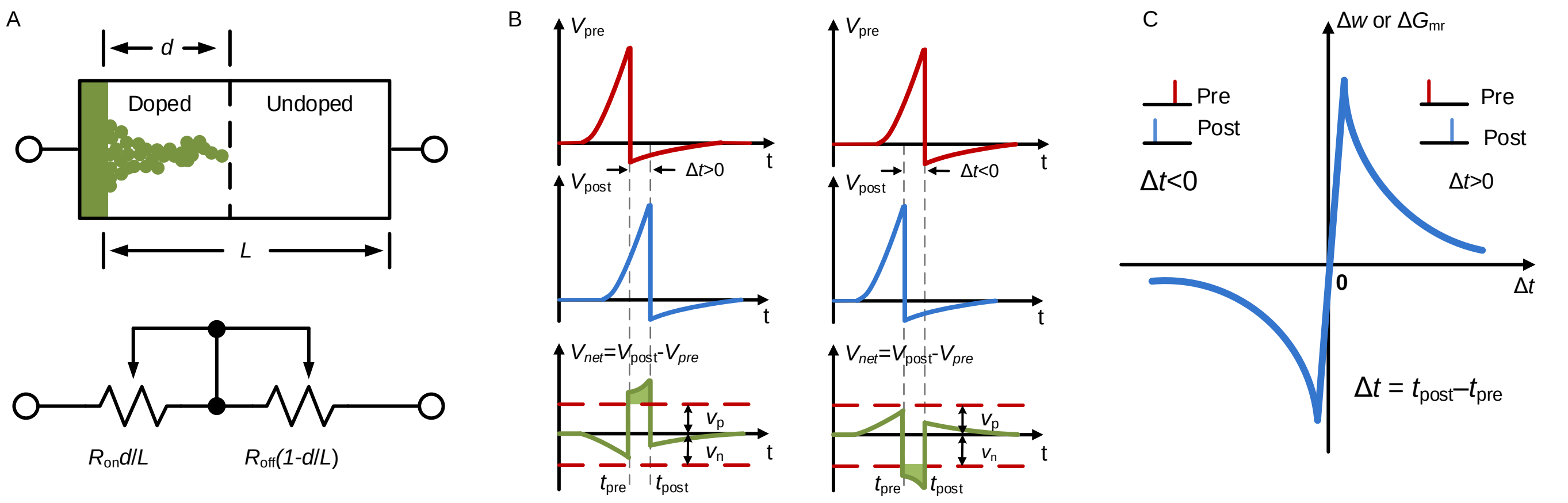

Fig. 2. (A) A thin-film memristor is modeled as two resistors in series, one is undoped with high resistance $R_{off}$ and the other is doped thus having low resistance $R_{on}$. To increase the depth of doped region, ions are forced into the film with a potential over programming threshold, $V_p$, on the two electrodes; conversely, to reduce the depth of doped region, ions are removed from the film with an opposite potential over erasing threshold $V_n$. (B) Pre- and post-synaptic spikes with relative arriving time $\Delta t$ produce a potential $V_{net} = V_{post} - V_{pre}$ over a synapse. $V_{net}$ over a threshold $V_p$ or $V_n$ leads into synaptic potentiation or depression, which for a memristor is equivalent to conductance increment and decrement respectively, caused by doping depth modulation. (C) A example of pairwise STDP learning window $\Delta w$ plotted as a function of $\Delta t$. Several nano-scale memristors demonstrated similar function with conductance change denoted as $\Delta G_m$.

two-terminal devices using various material systems [13]–[19], [38]–[42]. Fig. 2A schematically shows a highly simplified model of thin-film memristors, in which a memristor is composed of two resistors in series, one is un-doped with high resistance and the other is doped thus having low resistance. The total thickness of the film $L$ is separated into doped and un-doped regions, and the total resistance is the sum of the two regions. The average length of the doped region is taken as a state variable $d$. To increase the depth of the doped region, ions are forced into the film with a potential over the threshold $V_p$ across two electrodes; on the contrary, to reduce the depth of the doped region-, ions are removed from the film with an opposite potential which exceeds the erasing threshold $V_n$. This modulation of the doping depth allows the control of the conductance of a memristor. It should be noted that the above two-resistor model is a simple and convenient way of describing a memristor. In the dielectric region of a physical memristor device, the doping depth is typically represented by complex metallic filament structures. There exist a multitude of models that aim to correspond to the physics/chemistry behind the conductance change in memristors of various types [18], [19], [43], [44]. In this work, a much more sophisticated device model pertinent to physical memristors, from [44], was used for circuit simulation.

Several nano-scale memristors in literature have shown that their conductance modification characteristics are similar to the STDP rule [13]–[17], [45], and therefore act as ideal electrical synapses for brain-inspired computing. STDP states that the synaptic weight $w$ is modulated according to the relative timing of the pre- and post-synaptic neuron firing. As illustrated in Fig. 2B, a spike pair with the pre-synaptic spike arrives before the post-synaptic spike results in increasing the synaptic strength (or potentiation); a pre-synaptic spike after a post-synaptic spike results in decreasing the synaptic strength (or depression). Changes of the synaptic weight plotted as a function of the relative arrival timing of the post-synaptic spike with respect to the pre-synaptic spike is called the STDP function or learning window. A popular choice for the STDP function $\Delta w$ is shown in Eq. 1, and the corresponding plot is shown in Fig 2C

$$\Delta w = \begin{cases} A_+ e^{-\Delta t/\tau_+} & for\ \Delta t > 0 \\ A_- e^{\Delta t/\tau_-} & for\ \Delta t < 0 \end{cases} \tag{1}$$

A theoretical analysis in [33] illustrated a method to relate $\Delta w$ and memristor characteristics, by mapping the over-threshold portion of $V_{net}$ (the shaded area of the shaded regions in Fig 2B) to the change in memristance through an ideal memristor model. However, physical devices have complicated physical and/or electro-chemical mechanisms. Consequently, researchers typically plot a memristor's conductance $\Delta G_{mr}$ versus $\Delta t$ either from simulations or experimental results to show the STDP learning function.

Nano-scale memristors have shown low-energy consumption to change their states and very compact layout footprint [18], [19], [46]. Recent advances even reported these two merits in sub-pJ order [47], and 10-nm range [48] respectively. Thus, it is possible to yield a brain-inspired machine by cohesively packing millions of memristor synapses and thousands of CMOS neurons on a stamp-size silicon chip while consuming power density which is of the same order as a human brain (for a nominal 1kHz spiking rate).

### *B. Silicon Neuron*

Since neuromorphic engineering emerged in 1980s [49], several silicon neuron design styles have appeared in literature. These designs model certain aspects of biological neurons [50]–[57]. However, most of them focus on faithfully modeling the ionic channel dynamics in biological spiking neurons, and require the synapses to act as controlled current sources. As a result, they consume large silicon area, and therefore are not amenable for large-scale neuromorphic networks with a massive number of silicon neurons.

The emergence of nano-scale memristors has triggered a growing interest in integrating these devices with silicon neurons to realize novel neuromorphic systems [23]–[32]. In these systems, researchers have used bio-inspired leaky integrate-and-fire neuron (IFN) models as an alternative to the complex bio-mimetic neuron models to implement large networks of interconnected spiking neurons. The IFN model is a single-compartment model, wherein the entire cell is abstracted as a single membrane capacitance $C_m$ which sums each current $I_i(t)$ flowing into the neuron through the $i^{th}$ synapse, and a membrane resistance $R_m$ which causes passive leakage of a membrane current $V_m(t)$ / $R_m$ as

$$C_m \frac{dV_m}{dt} = \sum_i I_i(t) - \frac{V_m(t)}{R_m}. \quad (2)$$

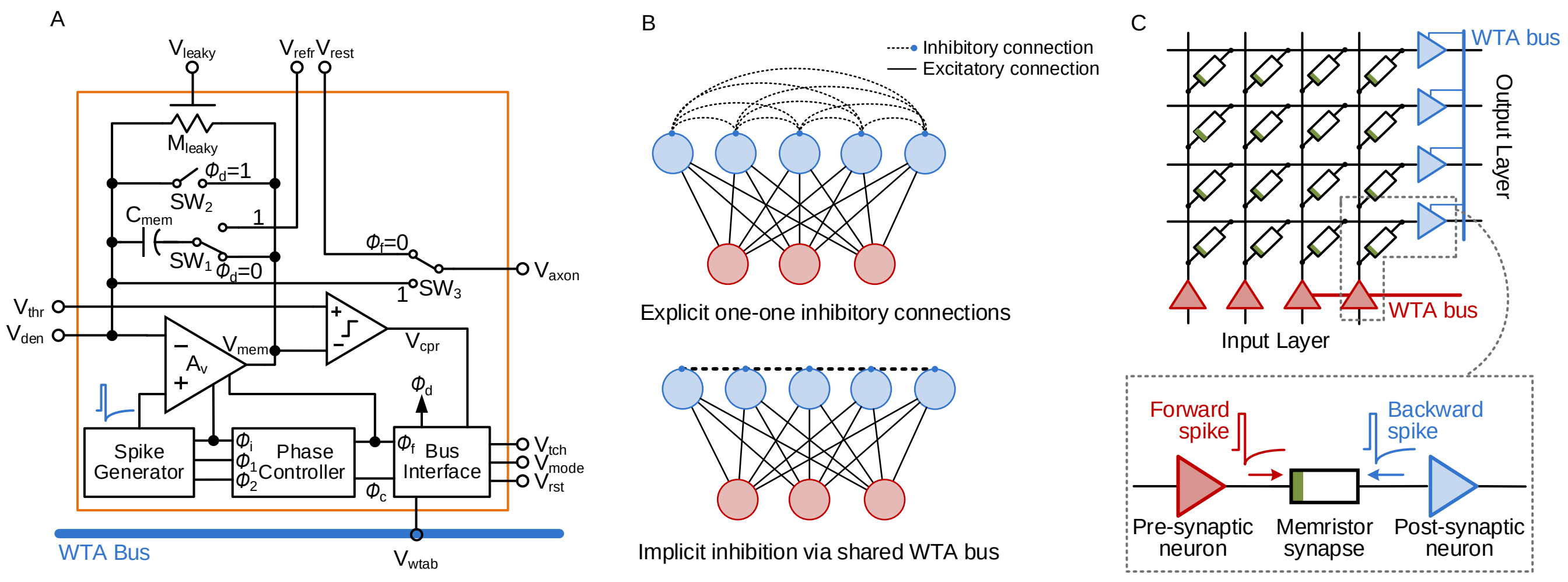

Fig. 3. (A) Diagram of the proposed leaky IFN. It includes integrate-and-fire, WTA interface, STDP-compatible spike generation, large current driving ability and dynamic powering in a compact circuit topology with a reconfigurable architecture based on a single OpAmp. (B) A competitive learning network uses explicit one-on-one inhibitory connections among competitive units; whereas the same function can be implemented with implicit inhibition on a shared WTA bus. (C) A layer of spiking neural network with memristor synapses organized in crossbar. Each input and output neuron pair is connected with a two-terminal memristor synapse. An STDP spike pair is used to update synaptic weight online without extra training circuitry. The WTA bus shared among output neurons enables the local competitive learning.

The IFN model captures the transient spiking behavior of the neuron with reasonable accuracy for use in learning while requiring a relative low number of transistors for its implementation. Currently, the IFNs used in memristor neuromorphic systems need either extra training circuitry attached to memristor synapses (thus eliminating most of the density advantages gained by using memristor synapses) or employ different waveforms for pre- and post-synaptic spikes (thus introducing undesirable circuit overhead which limits power and area budget of a large-scale neuromorphic system).

There have been a very few CMOS IFN designs attempting to address above problems in order to accommodate memristor synapses with in-situ synaptic plasticity ability. In [33], the authors proposed a reconfigurable IFN architecture which provided a current summing node to accommodate memristors. In [34], an architecture with a STDP-compatible spike generator was proposed, which enables online STDP by propagating same-shape spikes in both the forward and backward directions. In [35] a CMOS IFN with a current conveyor was designed to drive memristor as either an excitatory or an inhibitory synapse, and [36] shows the measurement results from a ferroelectric memristor. However, none of them can be directly employed to form a learning system because a decision making ability (e.g. competitive learning) was absent in these neurons. They require extra decision circuitry which may need a large silicon area and doesn't correspond to its biological counterparts. Moreover, these neurons don't provide an energy-efficient driving capability to interface with a large number of memristor synapses, which is generally desired in mimicking biological neural networks, e.g. a cerebellar Purkinje cell needs to form up to 200,000 synaptic connections [58], or for real-world pattern recognition applications, e.g. MNIST patterns have 784 pixels [59]. For instance, when a neuron drives 1000 memristor synapses, each of them having 1MΩ resistance, it requires 1mA current to sustain a 1V spike amplitude resulting in 1mW instantaneous power consumption. Therefore, a highly-scalable driver circuit solution for memristor synapses while avoiding large circuit overhead is truly desired [22].

A silicon neuron amenable to build large-scale brain-inspired neuromorphic system with massive memristor synapses should:

(1) Connect to a synapse at one terminal only;
(2) Sustain a fixed voltage across the synapse in the absence of spikes;
(3) Provide a current summing node to sense incoming spikes;
(4) Provide large current flowing into synapses when firing;
(5) Fire a suitable waveform to enable STDP in the synapse;
(6) Enable pattern learning through decision-making ability;
(7) Be compact and energy-efficient.

Fig. 3A shows the schematic of our proposed CMOS neuron that fulfills all of the above criteria. This circuit effectively combines an OpAmp-based integrator, an STDP-compatible spike generator, a WTA interface and a control circuit for reconfiguration. By employing tri-mode operation, it provides a unique port, $V_{den}$, to sum the incoming currents and to propagate post-synaptic spikes, and another port $V_{axon}$ to propagate pre-synaptic spikes. These two ports also sustain a fixed voltage $V_{refr}$ during integration and membrane capacitor discharge, while driving a specific STDP-compatible waveform with a large current to enable online synaptic plasticity in the large number of memristor synapses connected in parallel. Moreover, an inhibitive discharge mode with a shared WTA bus enables competitive learning among local neurons. All of these functions are assembled around a single CMOS OpAmp that is dynamically biased to supply large current only when driving the synapses while maintaining low power consumption during the rest of the time. Further, the neuron functions in a fully asynchronous manner consuming dynamic power only when computation is occurring. The details of the neuron circuit and its operation will be discussed in section III.

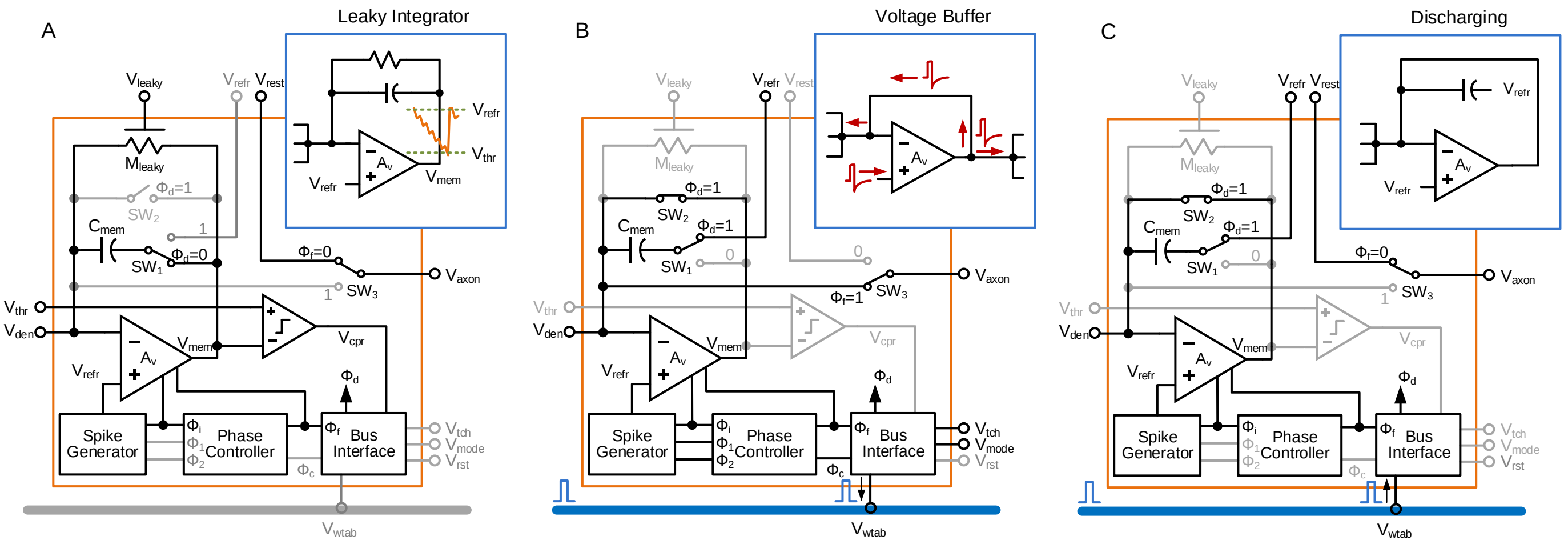

Fig. 4. Tri-mode operation of the proposed leaky integrate-and-fire neuron (A) Integration mode: The OpAmp is configured as a negative integrator to sum current on $C_{mem}$ causing the membrane potential $V_{mem}$ to move down until its crosses a firing threshold voltage $V_{thr}$. Without an input current, voltages at the two inputs of the OpAmp are held at $V_{refr}$. Post-synapses are disconnected from the neuron. (B) Firing mode: phase signals $\Phi_i$, $\Phi_f$, $\Phi_1$ and $\Phi_2$ control the spike generator to create a STDP-compatible spike $V_{spk}$ which is buffered and driven by the OpAmp. Then, the spike propagates in both backward and forward directions to pre-synapses and post-synapses respectively. The activation of either $V_{cpr}$ or $V_{tch}$ causes a firing event, which is also presented on the WTA bus by pulling-up the bus with $V_{wtab}$. (C) Inhibitive discharge mode: $\Phi_d$ is active to discharge the $C_{mem}$ when an active $V_{wtab}$ signal is detected on the WTA bus. The OpAmp is configured as a low-power buffer with $\Phi_i$ is active and $\Phi_f$ is inactive. Also, the neuron is isolated from the post-synapses.

### *C. Local Competitive Learning*

STDP enables online synaptic weight modification, but it doesn't automatically lead to network learning behavior. Conventional ANNs employ a gradient-based back-propagation algorithm to train a network. Although the same technique can be applied to SNNs as well [60], a gradient computation requires very sophisticated hardware and therefore is infeasible for a massively parallel system. In neuroscience studies, competitive learning has been observed and used to demonstrate synaptic plasticity directly together with STDP [12], [61]–[63], whereas no extra training circuitry is required. Competitive learning is also known as the winner-takes-all (WTA) algorithm whereby when a neuron fires, it inhibits its neighbors' from firing to prevent from changing their weight.

WTA uses a topology where an inhibit signal can be communicated to every other neuron in the network once it fires, at the same time, each neuron "listens" the inhibit signal from other neurons, as shown in Fig. 3B. However, such an explicit inhibition is resource hungry and difficult to scale-up in neuromorphic hardware, especially if the number of competing neuron units is large. Instead, an implicit inhibition with a bus-like operation is very efficient: several local neurons are connected to one shared bus together, and every neuron can monitor the bus status before its firing. In this scheme, a neuron is allowed to present an inhibitive signal only if there is no spike event on the shared bus; otherwise, it discharges and suppresses potential firing. The detailed circuit realization of the WTA bus will be discussed in section III.

It is worth noting that the proposed global reset mechanism differs from the dynamics of traditional neural networks, in which, typically, the firing of one neuron in a WTA network will either reduce the membrane potential (and thus spiking probability) of other neurons or prevent firing in a short time window. The implications to the computational aspects of the network dynamics with this global reset scheme can be investigated in further theoretical studies.

### *D. Crossbar Networks*

To build our proposed neuromorphic system, CMOS neurons and memristor synapses are organized in a crossbar network [64], [65], as shown in Fig. 3C. In this architecture, each input neuron is connected to another output neuron with a two terminal memristor to form a matrix-like connection for each crossbar layer. By cascading and/or stacking crossbars, a large-scale system can be constructed. Semiconductor technologies now offer vertical integration capability using through silicon via (TSV) for multiple chips and 3D packages [66].

As discussed, the proposed neuromorphic system architecture uses only two basic building blocks; a two-terminal memristor and a versatile CMOS neuron, which works in fully asynchronous manner. As they form a simple one-node contact, a large-scale neuromorphic system for brain-inspired computing can be potentially realized by spatially repeating and/or hierarchically stacking the proposed WTA circuit motif of neurons and crossbar synapses.

## III. The Design of CMOS Neuron

A silicon neuron is the most critical component needed to realize a neural network on a chip, while the synapses and crossbar structure are relatively simple in terms of architectural complexity. In our proposed neuron, the tri-mode operation, WTA bus, dynamic powering and STDP-compatible spike generation make up the key roles to realize a cohesive architecture.

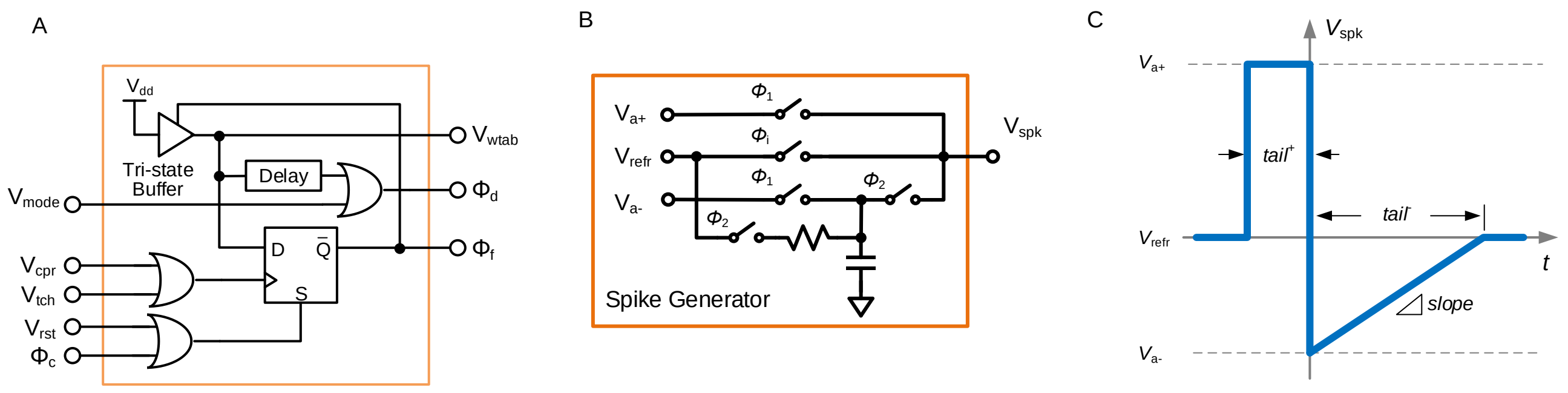

Fig. 5. (A) The proposed asynchronous WTA bus interface circuit. (B) STDP-compatible spike generator circuit. It produces (C) a spike with rectangular positive tail and ramping up negative tail. The spike shape is defined by parameters $V_{a+}$, $V_{a-}$, *tail*$^+$, *tail*$^-$ and *slope*.

### *A. Tri-mode Operation*

A spiking silicon neuron for competitive learning should perform three major functions: (1) current summing and integration, (2) firing when membrane potential crosses a threshold and driving resistive loads, and (3) providing an inhibitive discharge. These three functions are performed with a single OpAmp which is a key advantage of our neuron.

(1) The integration mode is shown in Fig. 4A. In this mode, switch $SW_1$ connects the "membrane" capacitor $C_{mem}$ with the output of the OpAmp, $SW_2$ is open, and $SW_3$ connects post-synapses to a resting voltage $V_{rest}$ which can be either equal to $V_{refr}$ or can be floated. $\Phi_d$ and $\Phi_f$ are asynchronous phase signals to control the switches. As the spike generator is designed to hold a voltage to the refractory potential $V_{refr}$ during the non-firing time, the OpAmp's positive port is set to $V_{refr}$. Under this configuration, the OpAmp realizes a leaky integrator; currents flowing from the pre-synapses are summed at $V_{den}$ and charge the capacitor $C_{mem}$ resulting in "membrane potential" $V_{mem}$, with the voltage leak-rate controlled by a triode transistor $M_{leaky}$. $V_{mem}$ moves down as more charge is stored on $C_{mem}$, and triggers a reconfiguration event of the neuron upon reaching the firing threshold $V_{thr}$.

(2) The firing mode is shown in Fig. 4B. In this mode, switch $SW_2$ is closed and the switch $SW_3$ bridges the OpAmp output to post-synapses. The OpAmp is now reconfigured as a voltage buffer. The STDP-compatible spike generator creates the required action potential waveform $V_{spk}$ and relays it to the positive port of the OpAmp. Then, both the pre-synapses and post-synapses are shorted to the buffer's output. The neuron propagates spikes in the backward direction from $V_{den}$ which is the same port of current summing. The pre-synaptic spikes are driven in the forward direction on $V_{axon}$ to the post-synapses. This firing-mode occurs either when the neuron wins the first-to-fire competition among the local neurons connected to a WTA bus, or during supervised learning. In the former scenario, the winning neuron presents a firing signal on the WTA bus noted as $V_{wtab}$, and forces other neurons on the same bus into "discharge mode". In the latter scenario, $V_{mode}$ indicates a supervised learning procedure and disables competition among the neurons. Then, with a teaching signal $V_{tch}$, the neuron is forced to fire a spike and drives it into pre-synapses, and consequently modulates the synaptic weights under the STDP learning rule. For stable operation, only one $V_{tch}$ of a neuron is active at a time in order to avoid conflict.

(3) The inhibitive discharge mode is shown in Fig. 4C. In this mode, switch $SW_1$ is closed, $SW_2$ connects $V_{refr}$ to discharge $C_{mem}$, and $SW_3$ is disconnected from the OpAmp output to isolate the neuron from the post-synapses.

### *B. Dynamic Powering*

The energy-efficiency of the neuron is tied to the above discussed tri-mode operation. For dynamic powering, a two-stage OpAmp is designed with the output stage split into a major branch and a minor branch. The major branch provides large current driving capability; while the minor low-power branch works with the first stage to provide the desired gain. Two complementary signals $\Phi_i$ and $\Phi_f$ are used to bias the OpAmp in low-power configuration by disabling the major branch during integration and discharging modes, while enabling it to drive large currents in the firing mode. In this work, we modified a compact folded-cascode topology [67] with an embedded split class-AB driver to realize a dynamically powered OpAmp.

### *C. WTA Bus Interface*

Fig. 5A shows a proposed WTA bus interface that can be embedded in the neuron with a compact implementation, and is amenable to scale-up. The bus interface works in an asynchronous manner. A tri-state buffer is employed to isolate the neuron output from the bus during the non-firing state, and a pulled-up bus when a neuron fires. During normal operation, the interface circuit monitors the bus status. A firing event presented as logic high on the bus activates $\Phi_d$ and forces the neuron to switch to the discharge mode. When a potential firing is triggered by either the comparator output $V_{cpr}$ or the supervised learning signal $V_{tch}$, the D-flip-flop (DFF) locks-in the instant bus state and passes it to $\Phi_f$. The logic low of $\Phi_f$, implying an existing firing event of another neuron, will consequently suppress neuron from firing; on the contrary, the logic high of $\Phi_f$ gives a green-light to switch the local neuron to the firing mode, and broadcasts an inhibitive signal via the shared bus. When the firing is finished, the DFF state is cleared.

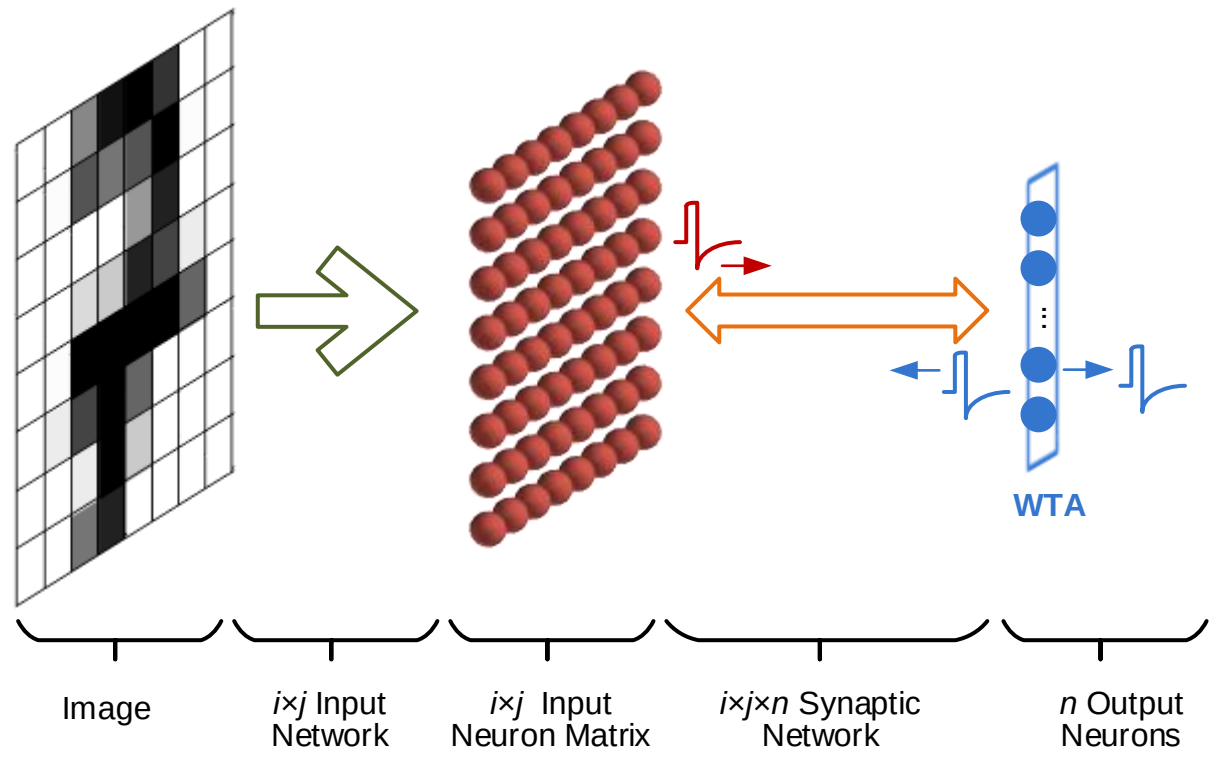

Fig. 6. A spiking neural system for the pattern recognition application of optical character recognition (OCR). Text images are sensed by an input neuron matrix with each pixel maps to a neuron, which converts it into spikes with spike rate proportional to the pixel darkness. All spikes from input neurons propagate through the memristor synapse network to the output neurons. Summing of input spikes causes a spike event from an output neuron with WTA competition. This spike from the output neuron acts as a decision signal and is used to update the synaptic weights with the STDP rule in training mode.

### D. STDP-Compatible Spike Generator

The shape of the action potential $V_{spk}$ strongly influences the STDP learning function. A biological-like STDP pulse with exponential rising edges is very difficult to realize in circuits. However, a bio-inspired STDP pulse can be achieved with a simpler action potential shape: a short narrow positive pulse of large amplitude followed by a longer slowly decreasing negative tail as plotted in Fig. 5C. This leads to a simple implementation, and yet realizes a STDP learning function similar to the biological counterpart [20]. The detailed spike generator circuit, shown in Fig. 5A, employs a voltage selector and RC charging circuit for the positive tail and the negative tail, respectively.

## IV. Pattern Recognition Application

As an important application of machine learning, optical character recognition (OCR) is widely used to demonstrate and evaluate pattern recognition performance. An electronic OCR system is designed to convert the images of printed text into computer-readable text to be used for electronic storage, pre-processing for machine learning, text-to-speech, and data mining, etc.

Fig. 6 illustrates a single-layer OCR system with the proposed architecture: the text image is read by an input sensory matrix where each pixel maps to a neuron and is converted into spikes. All spikes from input neurons propagate through a synaptic memristor network to the output neurons. Summing of the input spikes causes a spike from a winning output neuron under WTA competition, which then back-propagates and locally updates weights of the synapses via a STDP learning rule.

To effectively train this network, a supervised method is used. The teaching signal $V_{tch}$ is provided to the assigned output neuron as shown in Fig. 3A. The signal $V_{tch}$ forces the neuron to spike immediately after input pattern is received. Thus, the learning algorithm is tightly embedded in hardware in the proposed implementation.

In a trained network, test patterns can be classified without a teaching signal $V_{tch}$. Output neurons sum the currents flowing into them and fire according to the WTA competition to indicate the class of an input pattern. Such a pattern recognition system realizes real-time performance thanks to its straightforward event-driven parallel operation.

The proposed system is compatible with the spiking neural network model as described in [12], [61], [62]. Unsupervised learning of patterns can also be realized with the same circuit.

## V. Experimental Results

### A. Simulation Setup

The circuits were designed using the Cadence analog design environment and the simulations were carried out with the Spectre circuit simulator.

We employed a device model in [44] that has been matched to multiple physical memristors, and resistive random access memory characterization results.

The silicon neuron was realized with an IBM 180nm standard CMOS process. A two-stage OpAmp was used with folded-cascode topology for the first stage followed by a dynamically biased class-AB output stage. With an equivalent load of 1kΩ in parallel with 20pF, the OpAmp has 39 dB DC gain, 3V/μs slew rate and 5MHz unity-gain frequency in integration mode; and 60dB DC gain, 15MHz unit gain frequency and 15V/μs slew rate in firing mode. The STDP-compatible pulse generator circuit was designed with digital configurability to allow interfacing with a broad range of memristors. Such tunability may be also useful in the circuit implementation to compensate for the memristor parameter variations. For instance, spike parameters $V_a^+ = 140$mV, $V_a^- = 30$mV, $tail^+ = 1$μs and $tail^- = 3$μs were chosen for a device with $V_p = 0.16$V and $V_n = 0.15$V, where $V_a^+$ and $V_a^-$ were small enough to avoid perturbing the memristor, and large enough to create net potentials across the memristor with a potential above the memristor programming thresholds $V_p$ and $V_n$.

### B. CMOS Neuron Behaviors and STDP in Memristors

Functionality of the proposed neuron was first simulated in a small neural circuit with two memristor synapses connected between two input neurons (pre-synaptic neurons) and one output neuron (post-synaptic neuron) as shown in Fig. 7A.

Fig. 7B shows the integration and firing operations of the neuron and the STDP learning in the memristors. In this simulation, one of the pre-synaptic neurons was forced to spike regularly with output $V_{pre1}$ (solid line), while the other spikes randomly with output $V_{pre2}$ (dash line). The post-synaptic neuron summed the currents that were converted from $V_{pre1}$ and $V_{pre2}$ by the two synapses, and yielded $V_{mem}$. Post-synaptic spikes $V_{post}$ were generated once $V_{mem}$ crossed the firing threshold voltage $V_{thr} = 0.3$V. The bottom subplot shows potentiation and depression of the memristor synapses when a post-synaptic spike overlapped with the latest pre-synaptic spike, and created a net potential $V_a^+ + V_a^- = 170$mV over the memristors which was exceed their programming thresholds $V_p$ = 160mV or $V_n$ = 150mV. Quantitatively, a post/pre-synaptic spike pair with 1μs arriving time difference $\Delta t$ resulted in a 0.2μS conductance increase or decrease depending on late or earlier arrival of $V_{post}$ relative to $V_{pre}$ respectively. Fig. 7C summarizes the STDP learning in memristor conductance change $\Delta G_{mr}$ versus ±5μs range of $\Delta t$. The asymmetric curve shape with more depression peak value than potentiation was caused by the lower memristor negative threshold $V_n$ than $V_p$.

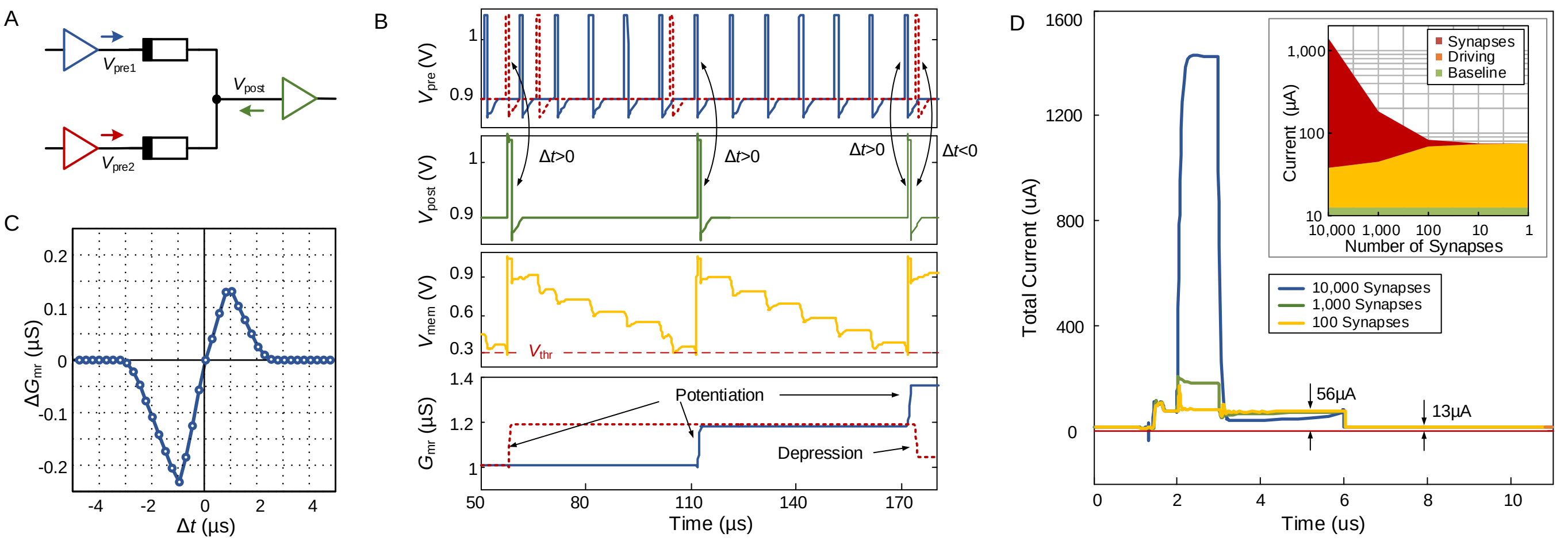

Fig. 7. (A) A small spiking neural network with two input neurons and one output neuron is used to demonstrate CMOS neuron operation. A memristor model in [44] was employed. (B) Neuron operation and STDP learning. Output neuron sums input current and yields the membrane potential $V_{mem}$. Post-synaptic spikes $V_{post}$ fired when $V_{mem}$ crossed $V_{th}$, and caused synaptic potentiation or depression, which depends on the relative arriving time with respect to the pre-synaptic spikes $V_{pre}$. (C) Simulated pairwise STDP learning window around 1μS conductance and 5μs relative time range. (D) Current proportional to synapse numbers was required to sustain spike voltage amplitudes for desired STDP learning in memristors, which causes large current being pulled when a large number of memristor are interfaced. Dynamic biasing based on dual-mode operation kept the neuron in very low power phase with only baseline (or static) current in integration mode, and extra current for output drive in firing mode. The embedded plot shows the current consumption breakdown versus the number of memristor synapses, assuming that the distribution of resistive states is tightly arranged around 1MΩ.

To evaluate energy-efficiency, the neurons were designed to have the capability to drive up to 10,000 memristor synapses with an assumption that the distribution of resistive states is tightly arranged around 1MΩ resistance. This yields a 100Ω equivalent resistive load. Fig. 7D shows the neuron consumed 13μA baseline current in the integration mode. When firing, the dynamically biased output stage consumed around 56μA current in the class-AB stage, and drove the remaining current to memristor synapses: a 1.4mA peak current for 10,000 memristor synapses sustained a spike voltage amplitude of 140mV. The current sunk by the synapses follows Ohm's law due to the nature of the memristor synapse as a resistive-type load. Insufficient current supplied to the memristors will cause a lower spike voltage amplitude that may fail STDP learning. Here, the widely used energy-efficiency figure-of-merit for silicon neuron, *pJ/spike/synapse*, becomes dependent on the resistance of synapses, and therefore, is not an appropriate descriptor of neuron's efficiency. Instead, the power efficiency $\eta$ during the maximum driving condition (at equivalent resistive load) should be used, i.e.

$$\eta = \frac{I_{\mathrm{mr}}}{I_{\mathrm{mr}}+I_{\mathrm{IFN}}}. \quad (3)$$

Here $I_{mr}$ is the current consumed by a memristor and $I_{IFN}$ is the current consumed by a silicon neuron. Our simulation demonstrated $\eta$ = 97% with 100 Ω for the selected memristor, and a baseline power consumption of 22μW with a 1.8V power supply voltage. This baseline power consumption doesn't change with the neuron's driving capability thanks to the tri-mode operation. As a comparison, a neuron without dynamical biasing consumes a 5-fold baseline current; a neuron based on dual-OpAmp architecture may consume a 10-fold static current. It should be noted these power consumption values are for a neuron design that targets a broad range of memristors, without optimizing for a specific device, and therefore have a significant room for improvement in power efficiency when designed for specific memristor characteristics.

### C. *Handwritten Digits Recognition*

We employed handwritten digits obtained from the UCI Machine Learning Repository [68] to demonstrate real-world pattern learning and classification with the proposed system. Fig. 8A shows the pattern examples in this dataset. These images include handwritten digits from a total of 43 individuals, 30 included the training set and a separate 13 to the test set. 32×32 bitmaps are divided into non-overlapping blocks of 4×4 and the number of 'on' pixels are counted in each block. This generates an input matrix of 8×8 where each element is an integer in the range of 0 to 15.

In our simulations, digits "0", "1", "2" and "7" were selected from the training dataset, in which there are 376, 389, 380 and 387 samples of each digit respectively. In the testing dataset, the samples number are 178, 182, 177 and 179, respectively. Samples in the testing dataset are different from the samples in the training dataset. These images were mapped onto an 8×8 sensory neuron matrix consists of 64 IFNs, and pixel values were converted into currents flowing to IFNs, with a threshold of seven or greater for "on" values used. This results in the input spike trains are shown in Fig. 8D. Each dot represents a spike and corresponds to an image pixel in binary form.

During the training phase, the training mode $V_{mode}$ signal was sent to the output neurons. Digit samples were presented to the system in their original sequence in the dataset. Corresponding labels were read into the simulator to activate the teaching signal $V_{tch}$ to the corresponding output neuron, and forced a post-synaptic spike $V_{post}$ at 1μs after each pattern was presented. All samples of the four digits in the training dataset were presented.

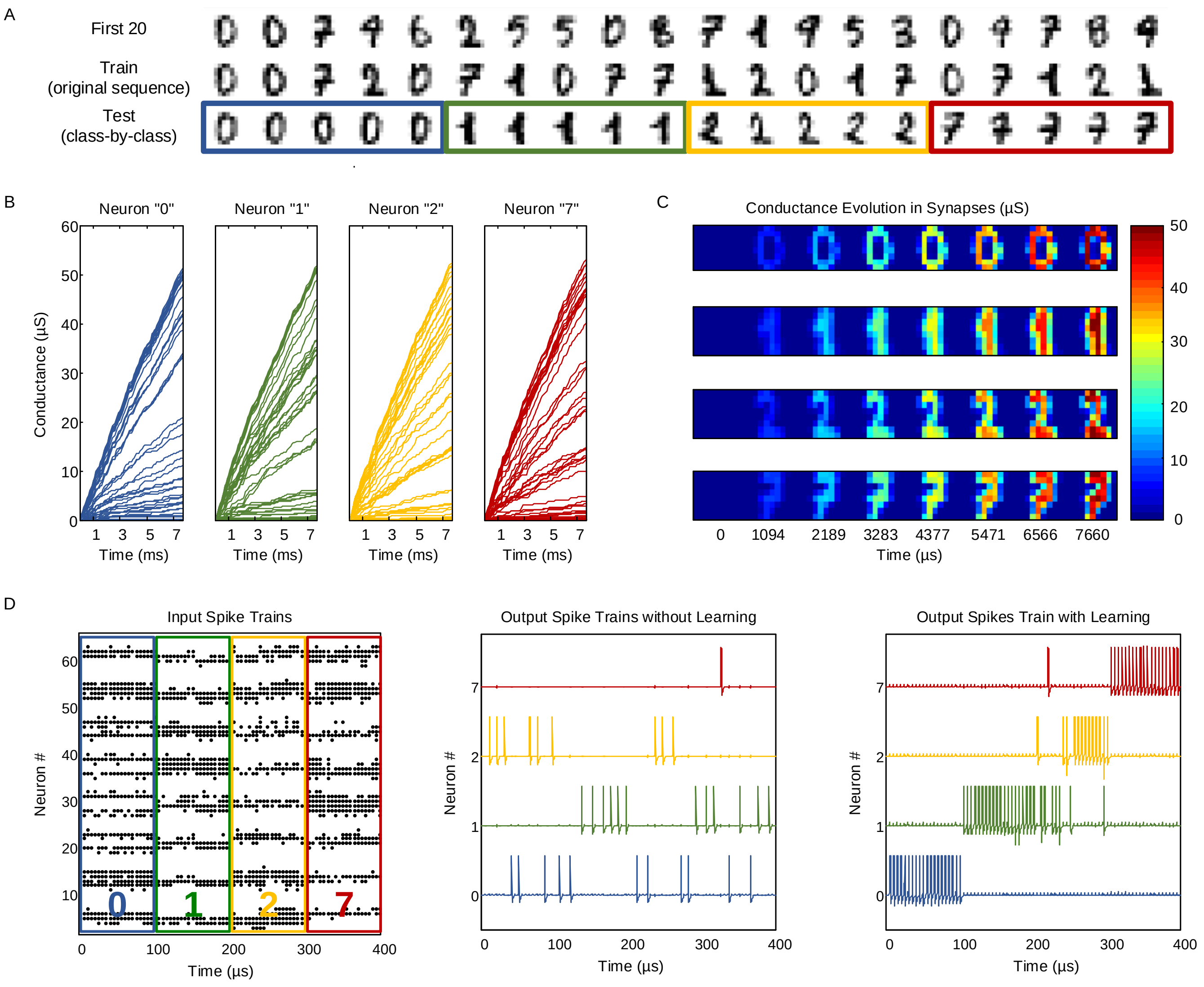

Fig. 8. (A) Examples of digits from UCI optical handwriting dataset. First line shows the first 20 digits images in the original training dataset, second line shows the first 20 digits samples used to train the network which is listed in the same sequence of original dataset, and the third line shows an examples of digits samples rearranged in class-by-class fashion used in testing but 5 samples for each digit. (B) Direct plot of memristor conductance learned in a circuit-level simulation with 4 output neurons during one epoch of training. (C) Conductance evolution rearranged as 8×8 bitmap. Before training, all synapses were initialized with a Gaussian random distributed conductance ($\mu$ = 8.5nS, σ = 4nS). After training, the maximum conductance is 53µS, and the minimum conductance is 6.6 nS. With the training moving on, the memristor network abstracted distinctive features of digits: loop of the digit "0", the vertical line of the "1", or the bone of "2" and "7". (D) Test results of the neural network with an input spike train composed of 20 samples for each digit and presented in class-by-class fashion. Without learning, a random synaptic network caused decision neurons spiking arbitrarily. After learning, each of these 4 output neurons is mostly selective to one of the 4 classes and spiking in the same class-by-class fashion of input.

Fig. 8B plots conductance changes in the memristor synapses connecting to each of the four output neurons. Before training, all synapses were initialized with Gaussian randomly distributed conductances ($\mu$ = 8.5nS, σ = 4nS). During training, their conductances were gradually increased and separated to different values, due to the STDP learning of the memristors. Because of computing resource restrictions on circuit-level simulations, we have limited the training demonstration to only one epoch here. However, the weights stabilize eventually after several epochs of training based on Matlab simulations as shown later using the IFN model of Eq. (2) instead of a transistor-level circuit.

Fig. 8C is a rearrangement of the conductance into an 8×8 bitmap with each pixel corresponding to an input image. It is remarkable that the synaptic networks abstracted several distinctive features of the digits: The loop of the digit "0", the vertical line of the "1", and the bone of "2" and "7".

Fig. 8D shows a testing simulation with 20 samples from each digit (out of four) and presented to the system for recognition in a class-by-class fashion. With an untrained synaptic network, the four output neurons responded to the inputs with random spiking. After training, each output neuron responds to the input patterns in the same class most of time showing clear selectivity, and only one neuron fired under the local competition rule.

Fig. 9A zooms into the details of currents and membrane voltages during testing. Due to the modulation of the synaptic network (causing different integration speeds), the total current flowing into the output neurons were separated; the neuron with the largest current ($I_0$) had its membrane voltage $V_{mem0}$ cross the firing threshold $V_{th}$ first winning the competition to fire first; whereas the current flowing into neuron "7" ($I_7$) was too small to make its $V_{mem7}$ reach the firing threshold. The other two neurons had their $V_{mem}$ reach the firing threshold, but their potential firing events were suppressed by the winner neuron. Membrane voltages of all neurons were reset by the WTA signal on the shared bus (not shown), and the actual circuit behavior introduced a 50ns delay from $V_{th}$ crossing to $V_{mem}$ resetting.

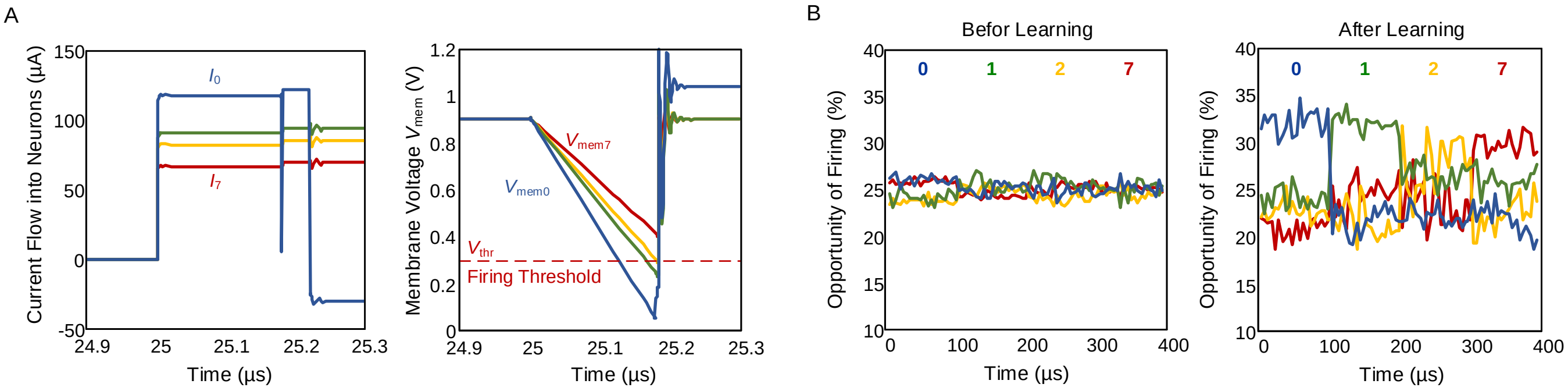

Fig. 9. (A) In a test case with one digit presented to the system, total current flowing into decision neurons were separated due to the modulation of synaptic network, which caused different integration speeds. The neuron with the largest input current $I_0$ had its membrane voltage $V_{mem0}$ cross the firing threshold $V_{th}$ first, and then won the competition of the race-to-fire first. (B) Firing opportunity and spike outputs of 4 output neurons for the spike input shown in Fig. 7D. All neurons have almost equal opportunities to spiking at the beginning. After learning, their spiking probabilities are modulated by their synaptic connections and distinguished. As result, a winner emerges.

To illustrate this competitive learning in another way, we define spiking 'opportunities' of the output neurons based on the total currents flowing into them,

$$p_n = \sum_i I_{n,i}(t) \ / \ \sum_n \sum_i I_{n,i}(t) \quad (4)$$

where $p_n$ is the relative spiking opportunity of the $n^{th}$ output neuron and $I_{n,i}$ is the current flowing into the $n^{th}$ output neuron by the $i^{th}$ input. With the same synaptic weights and the all $I_{n,i}$ equal, it follows that $p_n = 1/n$, which means the same chance to fire and no winner (for this reason, the synapses can't be initialized to all zero values. And such a condition doesn't exist in a real-world environment too). Once the synaptic weights are well modulated, they create different currents flowing into neurons. With a larger current, a neuron has the higher opportunity to spike in the same timeslot, which distinguishes the winner neuron from the others.

In this pattern recognition example, a 96% correction rate was achieved with the selected 4 digits. Matlab simulations with the IFN mathematical model show 83% correction rate with all 10 digits. These results are encouraging especially considering the system is a simple single-layer network, and no input encoding was applied. Applying symbolic patterns that were used in [24], [25], [28], [29], [69], [70], 100% correction rates were achieved simply because each pattern produced a unique synaptic network with their weights having exactly the same shape as the identical pattern of each class.

## VI. Discussion

The described CMOS spiking neuron architecture is generalized for memristor synapses. By selecting appropriate CMOS technology with sufficient supply voltage, online STDP learning can be achieved with the memristors, but not limited to, as reported in [39]–[41], [71]. However, the memristor in [13], with its $V_p = 1.5$V and $V_n = 0.5$V, would be difficult to fit into this architecture. With these threshold voltages, it is impossible to find a STDP pulse that can produce both potentiation and depression while not disturbing the memristor. In other words, for generalized STDP learning, assuming symmetric the pre- and post-synaptic spikes, a memristor is expected to have its thresholds satisfy the condition: $|V_p - V_n| < min(V_p, V_n)$.

In terms of energy-efficiency, an optimized design is the one with driving capability tailored according to the desired application and the memristor used. In the presented simulations, the neuron was tailored to support up to 1.5mA current in order to sustain $V_a^+ = 140$mV to a memristor network which has a peak average resistance around 93Ω. With MNIST patterns, each output neuron would have 784 input synaptic connections, thus the average resistive loading of these 784 synapses should be evaluated for both training and testing scenarios. The neuron driving capability is selected to sustain the least spike voltage amplitudes on the lowest equivalent resistive load while achieving the highest power efficiency. If the resistance of the memristor in its low resistance state (LRS) is 1kΩ and (say) 1% of the memristors are in their LRS, 7,840µA current is required to maintain a 1V spike voltage. For VGA (480×640 pixels) images, this number skyrockets to 32,700µA. It can be concluded that to implement low-power brain-inspired computing chip, the memristor synapses should have fairly high resistances (more than a MΩ) in their LRS, or a mechanism to isolate non-active synapses from the network during neurons' firing without large overheads becomes necessary.

On physical device side, a memristor passive crossbar architecture generally suffers from sneak paths (undesired paths parallel to the intended path for current sensing) [18], [65], [72]–[74]. The sneak-paths problem is caused by directly connecting resistive-type cells on sensing grid to the high-impedance terminations of the unselected lines. As stated in section II. B, a fixed voltage across a memristor is required for brain-inspired computing. Therefore, every path without a spike in the crossbar is tied to $V_{refr}$, and so the above discussed large current pouring into memristor networks becomes costly in terms of power consumption. Theoretically, a non-firing neuron could have a floating output thus reducing the current, but consequently sneak paths may bridge spiking neurons to other neurons and cause malfunction. So far, none of the existing solutions for sneak-paths work for memristor synapses, and thus further studies are required.

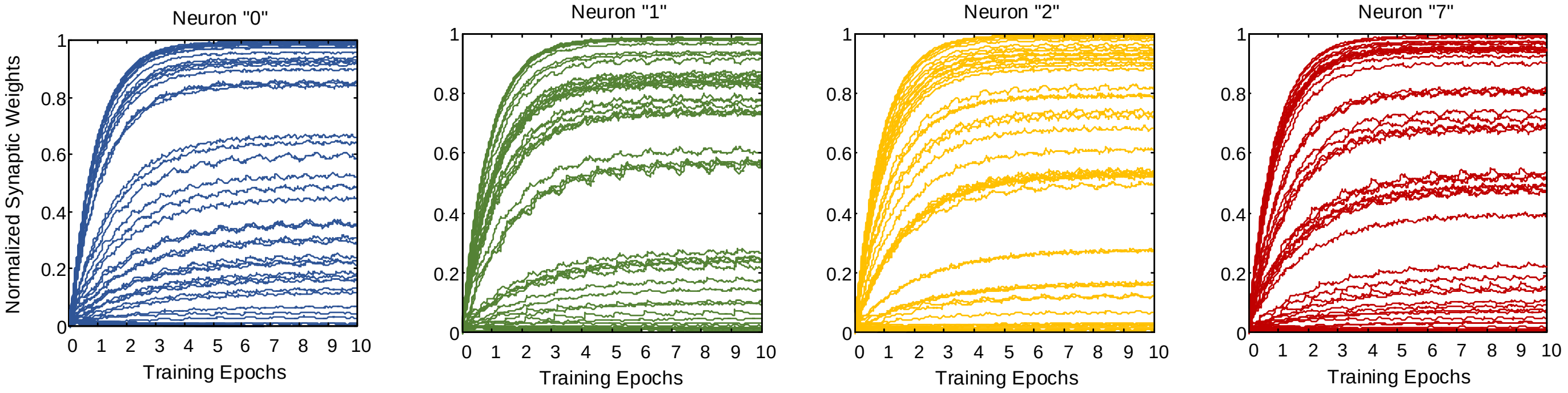

Fig. 10. Direct plot of memristor conductance learned in the behavioral Matlab simulation with the IFN mathematical model of Eq. (2). It shows the network optimally trains for the desired patterns and the weights eventually stabilize. This is expected as well if the circuit-level simulations were continued for several training intervals. The normalized synaptic weights in the Matlab simulation was initialized with a Gaussian randomly distributed ($\mu$ = 0.01, σ = 0.01) values.

Device variability is another challenge when using nano-scale memristors as synapses. Large variations in time and space of memristor synapses could cause unpredictable dynamics in the network, or simply fail to do learning. Although a spiking neural network offers some tolerance to device variation [75], the memristor threshold variations can easily fail network training especially when a low voltage spike is applied. There is a careful design trade-off between the low-voltage amplitudes of a spike required for energy-efficiency, and the high net potential margin over the memristor's characteristics required for reliable STDP learning. For instance, a memristor with $V_p$ = 160mV and $V_n$ = 150mV requires the spike voltage must higher than 80mV while a practical value typically in the range of 100 to 140mV to minimize the impact from device variations and spike noise. Some recent works have tried to address device variability by combining binary memristors to form a multi-level memristor cell for stochastic computing [32], [76]. Our proposed architecture works for stochastic computing as well, however, a stochastic firing mechanism is needed for the silicon neuron implementation instead of deterministic firing. Leveraging the stochastic behavior of nano-devices, a solution was proposed in [77] but its hardware realization feasibility still needs evaluation. Finally, it should be noted that the circuit-level simulations with faithful modeling of electrical behavior consumes significant amount of time as well as computing resources. Due to these restrictions, we limited the training demonstration to one epoch in the circuit-level simulations in shown this work. Based on the behavioral Matlab simulation results (see Fig. 10) with the IFN mathematical model of Eq. (2), the network optimally trains for the desired patterns and the weights eventually stabilize. This is expected if the circuit-level simulations were continued for several training intervals. Moreover, in our Matlab simulation, one has the flexibility to randomly initialize the weights. However, in a circuits approach, the memristors are expected to 'pre-formed' using a voltage pulse (or a photo-induced pre-forming step) which sets them in a high-resistance initial state. Therefore, the circuit simulations presented in this paper were initialized with all the memristors in their high-resistance state (low conductance) and then were potentiated to their final weights.

## VII. Conclusion

This paper describes a homogenous spiking neuromorphic system. It combines standard CMOS design of a novel silicon integrate-and-fire neuron with a memristor crossbar which can be realized in contemporary nano-scale semiconductor technology. This system naturally embeds localized online learning and computing by employing STDP learning in the memristor synapses with a winner-takes-all strategy among the local neurons. The CMOS neuron combines its circuit functions in a compact manner based on a single OpAmp, using a tri-mode operation. It also enables one-terminal connectivity between a neuron and a synapse, this fully exploits the synaptic density gain obtained by using memristor crossbar synapses. Supported by its reconfigurable architecture, a dynamic powering scheme allows the neuron to interface with a large number of memristor synapses without compromising energy-efficiency. Circuit simulations verified the functionality of the proposed neuron, and demonstrated an application of real-world pattern recognition with handwriting digits. In conclusion, the described system is homogenous, fully asynchronous, energy-efficient, and compact. Thus, it realizes a fundamental building block for a large-scale brain-inspired computing architecture.

## Acknowledgment

The authors thank the anonymous reviewers for their help in improving this work with their comments on the manuscript writing and results presentation. The authors also thank Dr. John Chiasson for his comments on the manuscript.